\RequirePackage{pdf14}
\documentclass{article}   
\usepackage{graphicx}
\usepackage{subcaption}
\usepackage{amsmath}
\usepackage{amssymb}
\usepackage{float}
\usepackage{url}
\usepackage{xcolor}
\usepackage{makecell}
\usepackage{cite}
\usepackage[colorinlistoftodos]{todonotes}
\newcommand{\eg}{\emph{e.g.}, }
\newcommand{\ie}{\emph{i.e.}, } 

\graphicspath{ {images/} }

\title{\LARGE \bf A Fast and Robust Place Recognition Approach for Stereo Visual Odometry Using LiDAR Descriptors}

\author{Jiawei Mo$^{1}$ and Junaed Sattar$^{2}$
\thanks{The authors are with the Department of Computer Science and Engineering, Minnesota Robotics Institute, University of Minnesota Twin Cities, Minneapolis, MN, USA.
{\tt\small \{$^{1}$moxxx066, $^{2}$junaed\} at umn.edu.}}
}
\date{\vspace{-5ex}}

\begin{document}

\maketitle
\thispagestyle{empty}
\pagestyle{empty}

\begin{abstract}
Place recognition is a core component of Simultaneous Localization and Mapping (SLAM) algorithms. Particularly in visual SLAM systems, previously-visited places are recognized by measuring the appearance similarity between images representing these locations. However, such approaches are sensitive to visual appearance change and also can be computationally expensive. In this paper, we propose an alternative approach adapting LiDAR descriptors for 3D points obtained from stereo-visual odometry for place recognition. 3D points are potentially more reliable than 2D visual cues (\eg 2D features) against environmental changes (\eg variable illumination) and this may benefit visual SLAM systems in long-term deployment scenarios. Stereo-visual odometry generates 3D points with an absolute scale, which enables us to use LiDAR descriptors for place recognition with high computational efficiency. Through extensive evaluations on standard benchmark datasets, we demonstrate the accuracy, efficiency, and robustness of using 3D points for place recognition over 2D methods. 
\end{abstract}

\section{Introduction}
\label{sec:introduction}
Visual SLAM (\textbf{vSLAM}) is an important capability for field robots, especially where GPS signal reception is weak or unavailable (\eg in urban or underwater settings). In these systems, visual odometry (VO) is used to build a local map and estimate ego-motion to assist in robot navigation. However, significant error can accumulate throughout the process, which causes odometry estimates to diverge from the correct path. Some form of a ``loop closure'' approach (\eg Bag-of-Words\cite{sivic2003video} \cite{galvez2012bags}) is required to recognize previously-visited places and bring non-local constraints into the system to get a globally consistent map and trajectory. Place recognition thus enables loop closures and improves VO accuracy.

Classical place recognition methods for vision-based systems usually rely on 2D images. Each location is represented by an image taken at that place. To determine the possibility of two locations being the same place, the similarity of their corresponding images is evaluated (see Sec. \ref{sec:related_work}). However, visual odometry methods provide additional information usable for place recognition purposes. The depth of points (\ie the distance of these points from the camera) on 2D images can be partially or fully recovered by monocular or multi-camera visual odometry, respectively. The 3D structure of the scene can potentially provide important information for place recognition; however, 2D place recognition methods ignore this. The 3D structure is more robust than 2D images in a dynamic environment (\eg under varying illumination). The motivation is also biological, as humans rely strongly on 3D structures for place recognition \cite{epstein1998cortical}.

\begin{figure}
\centering
\includegraphics[width=\textwidth]{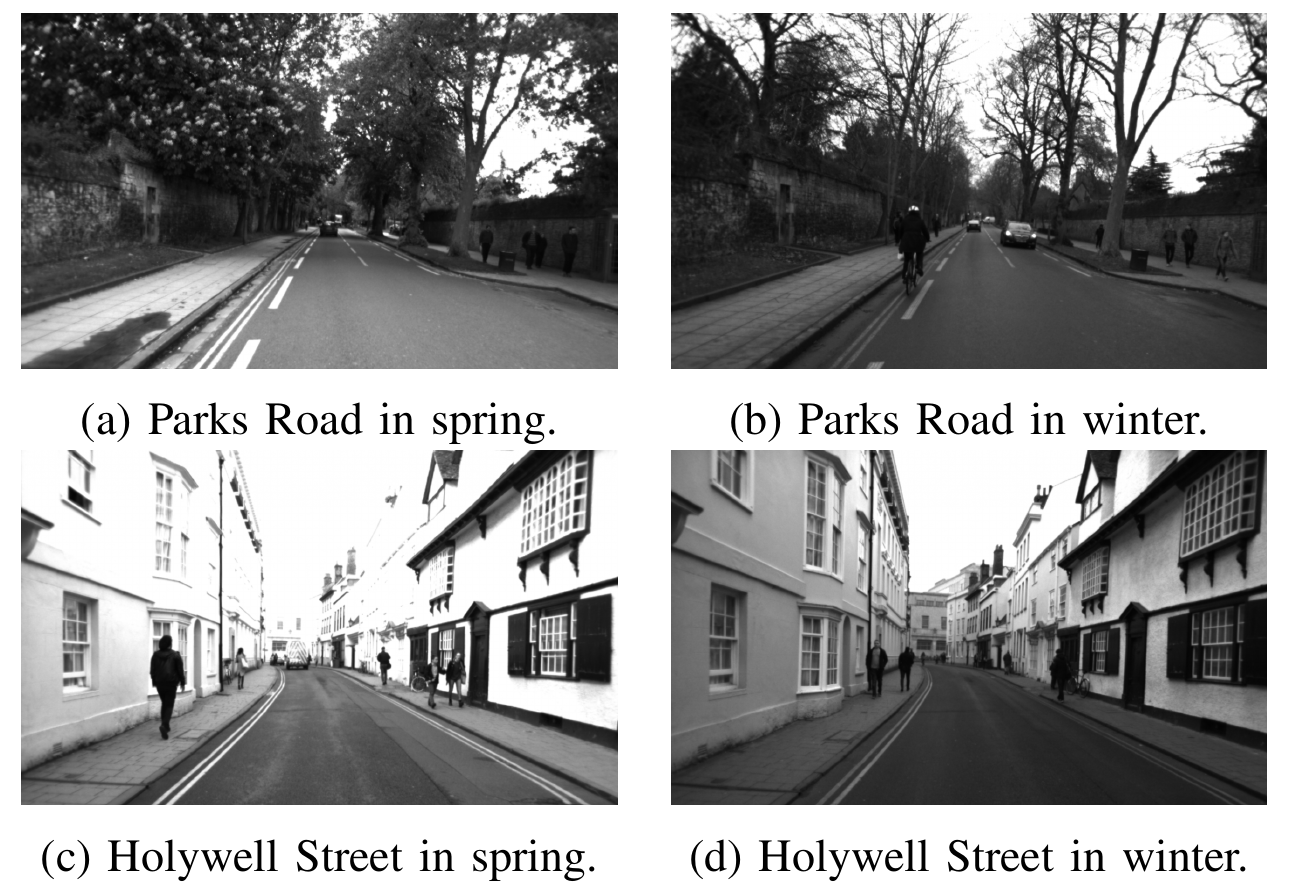}
\caption{Images from the RobotCar dataset in different seasons. Note the significant changes in appearance.}
\label{fig:robotcar}
\end{figure}

On the other hand, a rich body of literature exists on place recognition methods using 3D points from LiDAR (Light Detection and Ranging) sensors. LiDAR sensors scan the 3D structure of the environment rather than its visual appearance, making LiDAR-based place recognition more robust against environmental changes such as appearance and brightness (see Fig. \ref{fig:robotcar}). Another benefit of LiDAR methods is their high computational efficiency, and our evaluations demonstrate this when comparing 2D image-based and 3D LiDAR methods (see Sec. \ref{sec:experiments}).

In this work, we adapt LiDAR place recognition methods, in particular, LiDAR descriptors, into visual odometry systems for place recognition purposes. The goal is to enable accurate and robust place recognition in a computationally efficient way for a vision-based system in a dynamic environment. \textit{The proposed approach imitates a LiDAR range scan from 3D points generated by stereo-visual odometry, which enables us to adapt LiDAR descriptors}.

Several challenges must be overcome for applying LiDAR-based methods to vision-based systems. First, the 3D points generated by visual odometry are distributed in a frustum due to the much narrower field-of-view of cameras (excluding omnidirectional cameras) compared to most LiDAR sensors. The pose of the frustum changes with the camera, which is not desirable for place recognition. The second challenge is how to (and even if it is necessary to) adapt image intensity information into LiDAR-based methods, as such information is not available to LiDAR sensors. We address these challenges in this work. To the best of our knowledge, this is the first approach that uses global LiDAR descriptors for place recognition in vSLAM systems. The main contributions of this work, discussed in Sec. \ref{sec:methodology}, are as follows: 
\begin{itemize}
  \item Adapting global LiDAR descriptors to a vision-based system for place recognition,
  \item Achieving high accuracy and robustness against visual appearance changes,
  \item Achieving lower computational cost over existing approaches.
\end{itemize} 
    
We evaluate the proposed method on the KITTI dataset \cite{geiger2013vision} and the Oxford RobotCar dataset \cite{maddern20171}. We demonstrate the robustness of our method against drastic visual appearance changes across seasons as recorded in the RobotCar dataset, and show that it achieves higher accuracy and computational efficiency over existing methods. Further performance improvement is achieved by augmenting the LiDAR descriptor with image intensity information.

\section{Related Work}
\label{sec:related_work}
In the field of vSLAM, ORB-SLAM2 \cite{mur2017orb} is a recent development that demonstrates high accuracy and computational efficiency. In ORB-SLAM2, loop closure is detected by Bag-of-Words (BoW) using ORB features \cite{rublee2011orb}. A vocabulary tree is used in BoW to speed up feature matching and subsequent place queries. However, if the features are highly repetitive (\eg plants), BoW may fail; an example is given in Fig. \ref{fig:seq02}. Similarly, LSD-SLAM \cite{engel2014lsd} adopts FAB-MAP \cite{cummins2008fab} for place recognition. Other than BoW, Fisher vectors \cite{perronnin2007fisher} and VLAD \cite{jegou2010aggregating} also focus on 2D features. On the other hand, global image descriptors are also used to decide the similarity between images for place recognition. GIST \cite{oliva2006building} is one example which encodes spatial layout properties (spatial frequencies) of the scene. It exhibits high accuracy if the viewing angle does not significantly change. 

Recently, researchers adopted deep learning to place recognition and achieved impressive performance (\eg NetVLAD \cite{arandjelovic2016netvlad} and \cite{chen2017deep}). NetVLAD trained a convolutional neural network to extract learned features and proposed a generalized VLAD layer to describe the image automatically. Their accuracy is promising but their computational cost is usually high so that they are not widely used in real-time vSLAM systems. 

Neither BoW nor GIST is robust against visual appearance change, which is not ideal for long term (\eg from summer to winter) vSLAM applications, in addition to being computationally expensive. In ORB-SLAM2, place recognition runs in a separate execution thread to achieve real-time performance. Direct vSLAM systems (\eg \cite{engel2017direct}, \cite{forster2014svo}) have become popular in the past decade, which achieve higher performance in certain scenarios. Adapting BoW into direct vSLAM systems is challenging because features are not selected with the goal of being matched across frames. In LSD-SLAM mentioned above, an additional set of features are detected and matched separately, which are used specifically for place recognition, at a higher computational cost. In LDSO \cite{gao2018ldso}, the point selection strategy of its direct vSLAM system \cite{engel2017direct} is tuned to flavor features that can be matched across frames to enable BoW. Our proposed approach for place recognition, however, is more elegant for direct vSLAM systems if stereo cameras are available.
    
A number of 3D place recognition methods have been designed for RGB-D cameras or LiDAR sensors. RGB-D Mapping \cite{henry2012rgb} uses ICP \cite{besl1992method} to detect loop closure and RANSAC \cite{fischler1981random} to get an initial pose for ICP. For LiDAR, place recognition methods can be categorized into local descriptors and global descriptors. \textit{Local descriptors} use a \textbf{subset} of the points and describe them in a local neighborhood. Examples are Spin Image \cite{johnson1999using} and SHOT \cite{tombari2011combined}. Spin Image describes a keypoint by a histogram of points lying in each bin of a vertical cylinder centered at that keypoint. SHOT creates a sphere around a keypoint and describes that keypoint by the histogram of normal angles in each bin in the sphere. \textit{Global methods} describe the \textbf{entire set} of points. These methods can be more computationally efficient. Recent development includes NDT \cite{magnusson2009three}, M2DP \cite{he2016m2dp}, Scan Context \cite{kim2018scan}, and DELIGHT \cite{cop2018delight}. NDT classifies keypoints into line, plane, and sphere classes according to their neighborhoods. A histogram of these three classes is created to represent the point cloud. M2DP projects points onto multiple planes, and the histogram of point count in each bin on each projection plane is concatenated to get a signature of the point cloud. Scan Context aligns the point cloud to the vertical direction and represents it by the histogram of the maximal height of each bin on the horizontal plane. DELIGHT focuses on LiDAR intensity; the scan sphere is divided into 16 parts and the histogram of LiDAR intensity in each part is concatenated to represent the point cloud.
        
Cieslewski et. al. \cite{cieslewski2016point} looked into the possibility of using the 3D points triangulated from Structure-from-Motion or vSLAM for place recognition. They proposed the NBLD descriptor \cite{cieslewski2016point} for the 3D points from a vision-based system. A keypoint is described by its neighborhood points in a vertical cylinder. The point density of each bin in the cylinder is calculated and compared with neighborhoods to create a binary descriptor of that keypoint. Ye et. al. \cite{ye2017place} extended NBLD with a neural network. The vertical cylinder of NBLD is created in the same way; however, a neural network is trained to describe the cylinders, instead of calculating the point density. These are novel approaches in adopting point cloud descriptors into vision-based systems for place recognition.

In this work, we adapt global LiDAR descriptors into stereo-visual odometry for robust and efficient place recognition under visual appearance change. Direct vSLAM systems can easily adopt the proposed approach for place recognition without modifying their point selection strategy.
\section{Methodology}
\label{sec:methodology}
Similar to the idea of \cite{cieslewski2016point}, our method recognizes places based on the 3D points generated by visual odometry. The main difference is that the visual odometry in this work is running on stereo cameras. Specifically, we use SO-DSO~\cite{mo2019extending} as our stereo-visual odometry for its high accuracy and computational efficiency. To the best of our knowledge, SO-DSO is the only direct stereo-visual odometry that is robust to repetitive textures (Fig. \ref{fig:seq02}). We choose SO-DSO to demonstrate that the proposed method works for direct vSLAM systems. However, any multi-camera visual odometry is applicable here. Since the 3D points generated by stereo-visual odometry have an absolute scale, we can describe them efficiently using global LiDAR descriptors. The goal is high accuracy and computational efficiency, and robustness to environmental change.

\subsection{Point Cloud Preprocessing}
\label{sec:preprocess}
Due to the narrow field-of-view of the cameras, the 3D points generated by stereo-visual odometry are located in a frustum determined by the camera pose. If we apply a global descriptor directly inside the frustum, place recognition will be very sensitive to the viewing angle.

\begin{figure}[t]
    \centering
    \includegraphics[width=\textwidth]{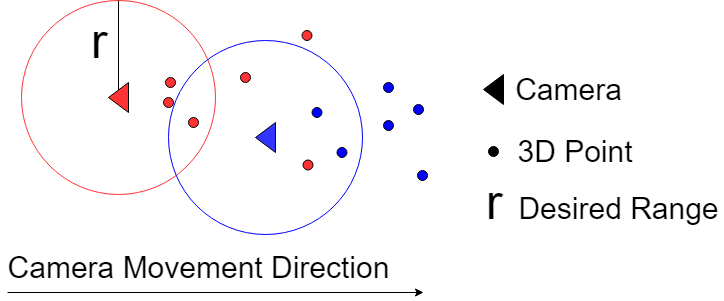}
    \caption{A demonstration of the points generated by visual odometry, with different colors representing different keyframes. 3D points are coming into the desired range as the camera moves.}
    \label{fig:pts_far}
\end{figure}
\begin{figure}[t]
    \centering
    \includegraphics[width=\textwidth]{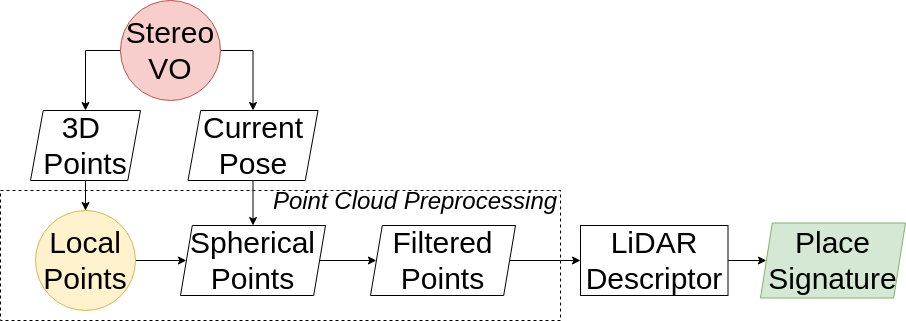}
    \caption{An overview of the proposed approach. The basis lies in the ``Point Cloud Preprocessing'' block, where 3D points obtained by stereo VO are used to imitate a LiDAR scan, so that efficient place recognition can be performed.}
    \label{fig:procedure}
\end{figure}

\begin{figure}
\centering
\includegraphics[width=\textwidth]{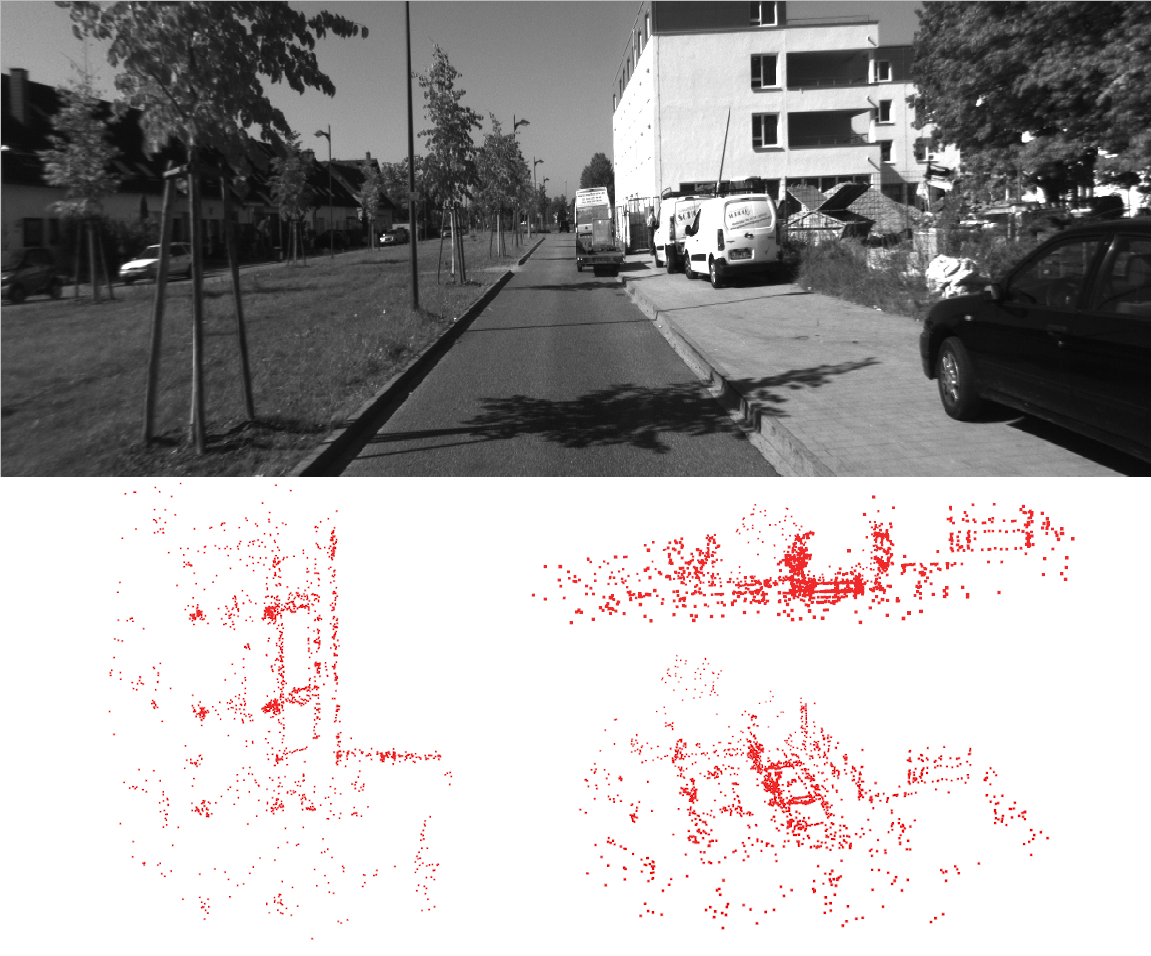}
\caption{Snapshots of an imitated LiDAR scan at the end of KITTI seq. 06.}
\label{fig:point_cloud}
\end{figure}

To solve this issue, we propose \textit{a simple but effective method that transforms 3D points from stereo-visual odometry in the frustum to an omnidirectional LiDAR-shaped (spherical) 3D point cloud}. The proposed method is illustrated in Fig. \ref{fig:procedure}. Stereo-visual odometry generates keyframes with camera poses and associated 3D points. We maintain what we refer to as a \textit{local points} list. For each incoming keyframe, we store all its \textit{3D points} into the list. To imitate a LiDAR scan for the current keyframe, we iterate through the \textit{local points} list: if the distance of the point is within the desired LiDAR range, we transform it to the current keyframe coordinate by \textit{current pose}, then put it into the \textit{spherical points} list. Here we assume the camera motion is predominantly in the forward direction so that we continuously have points coming into the desired range to compensate for points leaving the range, as illustrated in Fig. \ref{fig:pts_far}. The \textit{spherical points} may contain duplicate points; for robustness and computational efficiency, we filter them to get the final \textit{filtered points}. An example of an imitated LiDAR scan is given in Fig. \ref{fig:point_cloud}.

Caching local points enables us to imitate LiDAR scans with denser omnidirectional points. Since visual odometry generates locally accurate camera poses and 3D points, concatenating 3D points transformed from multiple nearby keyframes to imitate a LiDAR scan is feasible.

\subsection{Point Cloud Description}
\begin{figure}[t]
    \centering
    \includegraphics[width=\textwidth]{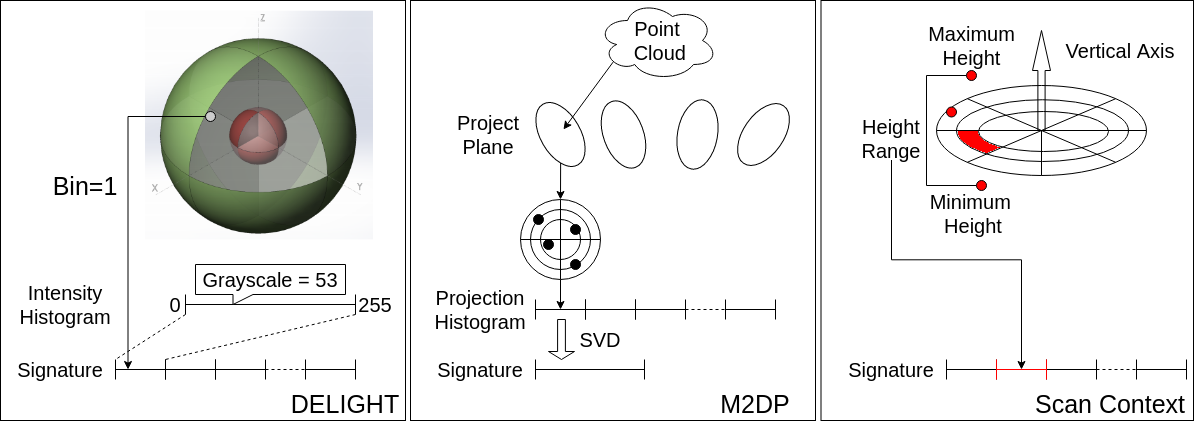}
    \caption{Demonstration of LiDAR descriptors used in this work: DELIGHT, M2DP, and Scan Context.}
    \label{fig:3d_desc}
\end{figure}

The next step in the proposed method is to describe the \textit{filtered points} and get a \textit{place signature} for the keyframe, for which we rely on global LiDAR descriptors. This is preferable for two reasons: the first is for its computational efficiency when describing and matching the point clouds; the second reason is due to the fact that the point clouds we have are generated by visual odometry, they are not as consistent and dense as the ones from a LiDAR. Many local descriptors, such as Spin Image, depend on the surface normal, for which dense point clouds are required, which would be problematic in this case. We choose DELIGHT~\cite{cop2018delight}, M2DP~\cite{he2016m2dp}, and Scan Context~\cite{kim2018scan} as our global descriptors since they are state-of-the-art LiDAR descriptors for place recognition that are robust to sparse and inconsistent point clouds. The high-level ideas of them are illustrated in Fig. \ref{fig:3d_desc}.

\paragraph*{DELIGHT}
DELIGHT operates on LiDAR intensities. The LiDAR scan sphere is divided into 16 bins by radius, azimuth, and elevation. Each bin is described by the histogram of LiDAR intensities inside, which are concatenated to form the signature representing the entire LiDAR scan. To make the descriptor less sensitive to rotation and translation, the raw LiDAR scan is aligned to a reference frame obtained by Principal Components Analysis (PCA)~\cite{tombari2010unique}. As discussed in~\cite{cop2018delight}, there are four versions of the signatures due to the ambiguity of PCA.

Analogous to LiDAR scan intensities, the 3D points from the visual odometry have grayscale intensities from the images. We simply replace LiDAR intensities with grayscale intensities and adapt the DELIGHT descriptor into our system. Each histogram is composed of 256 bins since the grayscale intensity ranges from 0 to 255. Although DELIGHT does not use 3D structural information, we include it in this work to contrast it against M2DP and Scan Context as both these methods use this information. This is done to highlight the value of 3D structures for robust place recognition against visual appearance change. 

\paragraph*{M2DP}
\label{sec:m2dp}
M2DP is a global descriptor that demonstrates high accuracy and efficiency. The point cloud is projected onto multiple planes, and each plane is separated into individual bins by radius and azimuth. The distributions of the projection onto bins are concatenated to form a signature for the point cloud. For computational and memory efficiency, singular-value decomposition (SVD) is used to compress the signature. As in DELIGHT, PCA is used to align the point cloud. 

In this work, we augment the M2DP descriptor using grayscale intensity from the visual odometry. Specifically, when projecting the point cloud onto each plane, we not only count the number of points projected onto each bin, but also calculate the average grayscale intensity. Therefore, we have two types of signatures (namely the point count signature and the intensity signature) for each place. To make the intensity signature less sensitive to illumination, we binarize the intensity by comparing it to the global average intensity. The intuition is to highlight the bright bins. To improve accuracy, we adopt the four versions of signature from DELIGHT. We include M2DP in our method as it utilizes 3D structure for place recognition and is not limited to urban scenarios like Scan Context.

\paragraph*{Scan Context}
Scan Context is a straightforward yet effective descriptor designed for LiDAR scans obtained in urban areas. The LiDAR scan is aligned with respect to the gravitational axis which is measured externally (\eg with an IMU). Then the horizontal circle plane is separated into multiple bins by radius and azimuth. In each bin, the maximum height is found and concatenated to form a signature for the current place. 

To fit Scan Context into our system, we make the following modifications. First, since we want to avoid using additional sensors, we adopt the PCA method used in DELIGHT and M2DP to align the point cloud. Second, due to the PCA ambiguity, we replace maximum height with height range (maximum height - minimum height). Lastly, we generate the intensity signature as in the modified M2DP. Scan Context is the most efficient and accurate among the three descriptors, as shown in experimental evaluations (Sec.~\ref{sec:experiments}).

\subsection{Place Recognition}
Based on the place signatures, we are able to determine the similarity between places. We generate a \textbf{difference matrix} by calculating the signature distance from each query place to every place in the reference database. For DELIGHT and M2DP, we take the shortest distance from the query signature to all four possible signatures of the reference place. For Scan Context, even though we can use the same approach as for DELIGHT and M2DP, we compare against all possible yaw angles (as the original Scan Context) for maximal accuracy. The distance of DELIGHT is based on the chi-squared test as described in \cite{cop2018delight}. For M2DP and Scan Context, the distance is simply the Euclidean distance between normalized (L2-norm=1) signatures. As we have two types of signatures (structure signature and intensity signature), we get two individual difference matrices $\mathbf{D_{s}}$ and $\mathbf{D_{i}}$. We fuse them in Eq. \ref{eq:diff_fuse} by normalizing (mean=0, std=1) each row (representing each query) and adding them with a relative weight $w_s$: 

\begin{equation}
    \mathbf{D_{fused}} = w_s\cdot N_{row}(\mathbf{D_{s}}) + N_{row}(\mathbf{D_{i}})
\label{eq:diff_fuse}
\end{equation}

With the difference matrix, each query place (row) is matched to a reference place with the smallest difference value among all candidates (along the row). 
\section{Experimental Evaluation}
\label{sec:experiments}
 To evaluate the proposed method, we compare the results internally among DELIGHT, M2DP, and Scan Context, as well as externally to BoW, GIST, and \cite{ye2017place}, as real-time performance at a low computational cost is a key aspect of our work. 

\subsection*{Implementation}
We use the fast (but less accurate) setting ($\leq800$ active points, $\leq6$ frames in optimization) of SO-DSO to estimate camera poses and generate 3D points. We found that the fast setting is good enough for our purpose and introducing more points decreases the computational efficiency. The outputs of SO-DSO (poses and points) are used by the proposed methods for place recognition.

The authors of M2DP have published their Matlab code, which we re-implement using C++. We similarly re-implement DELIGHT and Scan Context. When preprocessing the point cloud as discussed in Sec.~\ref{sec:preprocess}, we set the LiDAR range $r=45.0m$. For DELIGHT and M2DP, spherical points are filtered in polar coordinate with 1-degree angular resolution. We keep the closest point along each ray originating from the polar center. For Scan Context, however, we filter points in the Cartesian coordinate with $1.5m\times0.75m\times1.5m$ resolution. We switch to the Cartesian coordinate and assign a higher resolution along the vertical axis because Scan Context focuses on height. For the KITTI dataset \cite{geiger2013vision}, there are $2706.2$ ($1610.6$) points on average in each \textit{filtered points} using polar (Cartesian) filtering. For DELIGHT, the radius of the inner/outer sphere is set to $10$/$45$ meters, respectively. The parameters of M2DP and Scan Context are set to default values. For more details, our implementations are available online\footnote{https://github.com/IRVLab/so\_dso\_place\_recognition}. When fusing difference matrices in Eq. \ref{eq:diff_fuse}, we set a higher weight $w_s=2$ to structure because structure is more reliable than grayscale intensity.

The implementation of BoW comes from ORB-SLAM2; the implementation of GIST is available online\footnote{http://lear.inrialpes.fr/software}. 

\subsection*{Evaluation}
For place recognition accuracy, we focus on the area under the precision-recall curve (\textbf{AUC}) and the maximal recall at 100\% precision (no false positives, \ie no errors) as two indices. Larger AUC means more places are recognized with fewer errors. To some extent, AUC reflects the discrimination power of a place recognition algorithm. Larger maximal recall at 100\% precision indicates that more places are recognized before making any mistakes, it is important because a single false positive might significantly affect the accuracy of the entire SLAM algorithm. In the ideal case, both AUC and maximal recall should be 1. Other than accuracy, computational efficiency is also evaluated. We take apart each algorithm and compare the components individually. The accuracy and efficiency are evaluated on KITTI dataset \cite{geiger2013vision} and Oxford RobotCar Dataset~\cite{maddern20171}. 

\subsection*{KITTI Dataset} 
The KITTI dataset is one of the most influential datasets for benchmarking autonomous driving research. The odometry dataset comes with 22 stereo sequences. However, only the first 11 sequences have ground-truth publicly available. Among them, sequences \{00, 02, 05, 06, 07\} have loop closure segments, which are used in this section. 

\paragraph*{Accuracy}
\begin{table}
\centering
\tabcolsep=0.06cm
\begin{tabular}{|l|l|l|l||l|l|}
\hline
Method & DELI. & M2DP & S.C. & BoW & GIST \\
\hline
Seq. 00 & 0.754 0.616 & 0.639 0.191 & 0.733 0.599 & \textbf{0.893} \textbf{0.788} & 0.841 0.774 \\
\hline
Seq. 02 & 0.463 0.253 & 0.488 0.053 & 0.555 0.440 & 0.011 0.012 & \textbf{0.613} \textbf{0.597} \\
\hline
Seq. 05 & 0.622 0.483 & 0.522 0.062 & 0.653 0.566 & \textbf{0.867} \textbf{0.809} & 0.756 0.659 \\
\hline
Seq. 06 & 0.916 0.531 & 0.946 0.671 & 0.897 0.679 & \textbf{0.968} \textbf{0.963} & 0.925 0.729 \\
\hline
Seq. 07 & 0.000 0.000 & 0.000 0.000 & 0.000 0.000 & \textbf{0.713} \textbf{0.627} & 0.350 0.149 \\
\hline
\end{tabular}
\caption{AUC (first number) and maximal recall at 100\% precision (second number) on KITTI dataset.}
\label{tb:kitti-accuracy}
\end{table}

\begin{figure}
\centering
\includegraphics[width=\textwidth]{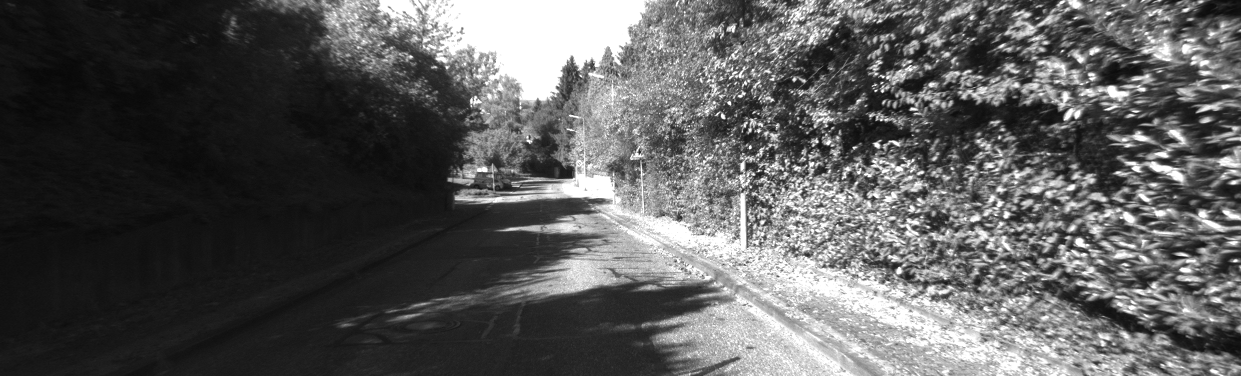}
\caption{Snapshot of KITTI seq. 02 at a revisited place.}
\label{fig:seq02}
\end{figure}

\begin{figure}
\centering
\includegraphics[width=\textwidth]{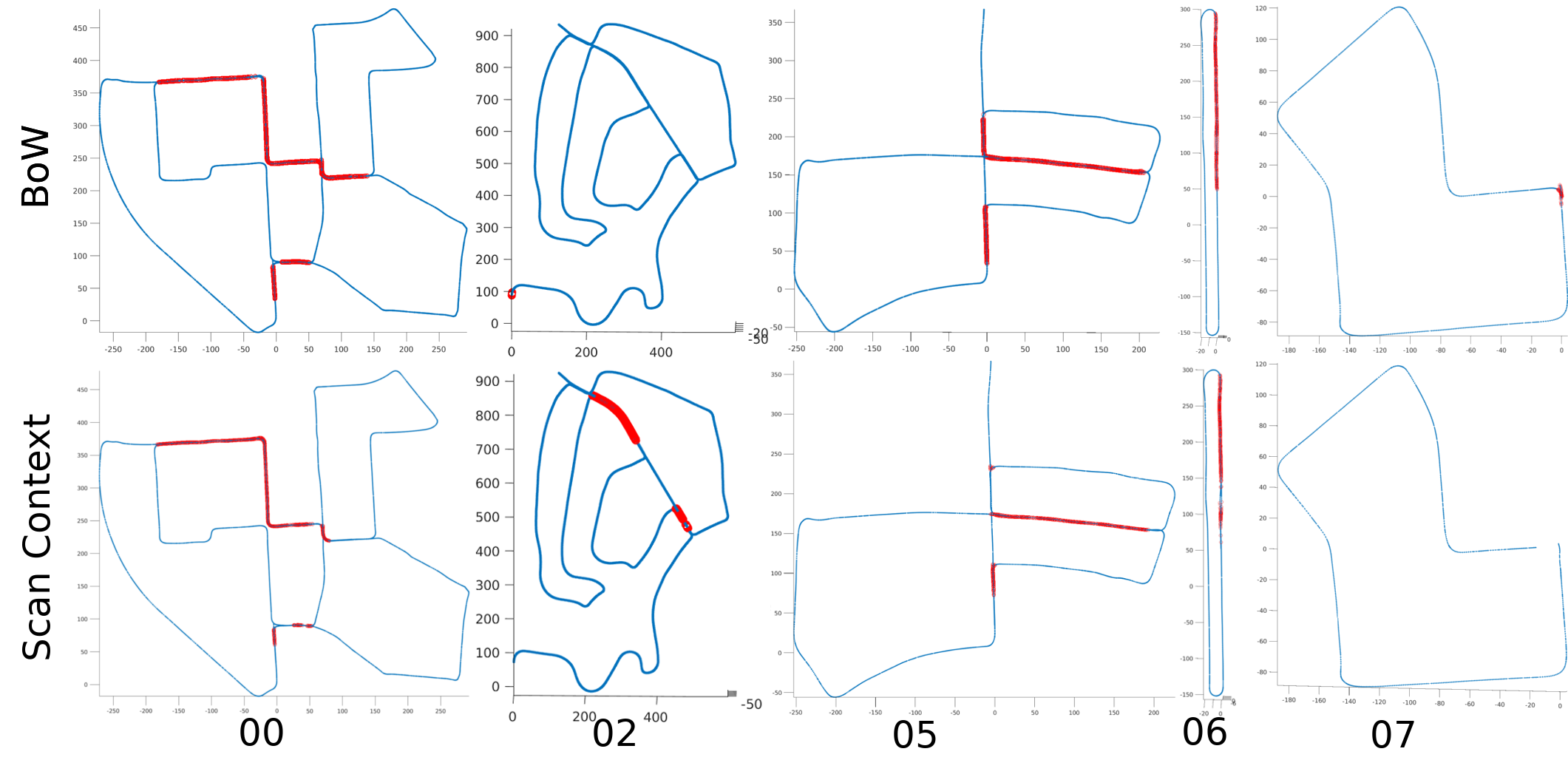}
\label{fig:kitti}
\caption{Places recognized (marked as red) by BoW and by Scan Context on KITTI dataset at 100\% precision.}
\label{fig:kitti-plot}
\end{figure}

When computing precision and recall, two places are considered to be revisited places if their distance is smaller than 10 meters, which is relatively small as the distance between trajectories can be small (\eg sequence 06 in Fig. \ref{fig:kitti-plot}).

Table \ref{tb:kitti-accuracy} reports the accuracy of each algorithm. BoW achieves the best accuracy in all sequences other than sequence 02. Each sequence in the KITTI dataset is recorded continuously in a short period of time, there is not much visual appearance change. Hence, feature matching is robust and BoW works perfectly. For sequence 02, the BoW approach fails to recognize places because the revisited places are occupied with repetitive textures (\ie plants in Fig. \ref{fig:seq02}), for which feature matching is unreliable. The accuracy of the adapted LiDAR approaches is not as good as BoW or GIST but their AUCs are still fairly high on sequences \{00, 02, 05, 06\}. Scan Context achieves the best overall accuracy among the adapted LiDAR approaches. Sequence 07 is special because there is only a small segment of loop closure when the vehicle comes back to the starting place. Thus, the accuracy of BoW and GIST is not very reliable. None of the proposed 3D methods detects any loop because there is not enough overlapped trajectory to accumulate 3D points. 

Fig. \ref{fig:kitti-plot} plots the loops detected by BoW and Scan Context. Scan Context fails to recognize places in short segments. This is because the proposed method needs enough overlapped trajectory to accumulate 3D points. This is a limitation of the proposed method. However, it recognized places with repetitive textures in sequence 2. Nevertheless, we showed that our implementation of each algorithm works on the KITTI dataset (at least for BoW and GIST; we will further validate the proposed 3D approaches using the RobotCar dataset). 

\paragraph*{Efficiency}
\begin{table}
\centering
\tabcolsep=0.06cm
\begin{tabular}{|l|l|l|l||l|l|}
\hline
Method & DELI. & M2DP & S.C. & BoW & GIST \\
\hline
Imitate LiDAR Scan (c++) & 1.151 & 1.204 & \textbf{0.692} & - & - \\
\hline
Desc. extraction (c++) & \textbf{0.082} & 46.10 & 0.123 & 37.41 & 160.0 \\
\hline
Query descriptor (Matlab) & 103.2 & 3.418 & 7.334 & 115.0 & \textbf{1.106} \\
\hline
Total & 104.4 & 50.72 & \textbf{8.149} & 152.4 & 161.1 \\
\hline
\end{tabular}
\caption{Run time analysis in milliseconds.}
\label{tb:run-time}
\end{table}

Table~\ref{tb:run-time} reports the run-time required to query a place in the database. The test platform is based on an Intel i7-6700 with 16GB of RAM. The run-time is calculated in sequence 06.  

Point filtering in Scan Context is slightly faster because of the simple Cartesian coordinate filtering. For descriptor generation, DELIGHT is the fastest due to its straightforward mechanism. Scan Context is the second-fastest because calculating the height range is also efficient. M2DP is the slowest one among all the adapted 3D methods. The entire set of points is projected onto multiple planes, followed by SVD compression, which is computationally expensive. For BoW and GIST, their high accuracy on the KITTI dataset is achieved at a high computational cost. For place query, GIST is the quickest since it simply calculates the Euclidean distance between two descriptors. This is followed by M2DP, for which we have calculated all four descriptors of each place, the distance between two places is simply the smallest Euclidean distance. Scan Contest is slightly slower because the query descriptor is matched against all possible yaws. DELIGHT and BoW are much slower. The chi-squared test in DELIGHT is computationally expensive. For Bow, we extract $4000$ features for each image and use a vocabulary tree with about 1 million nodes. Hence, both descriptor extraction and matching of BoW are computationally expensive.

Scan Context achieves the highest overall efficiency that can run in real-time in most vSLAM systems. Although BoW and GIST achieve higher accuracy, they are much slower than Scan Context.

\subsection*{RobotCar Dataset}
\begin{table}
\centering
\tabcolsep=0.06cm
\begin{tabular}{|l|l|l|l|l|l|l|l|l|l|l|}
\hline
Tests & \makecell{Spr.\\Spr.} & \makecell{Spr.\\Sum.} & \makecell{Spr.\\Fall} & \makecell{Spr.\\Win.} & \makecell{Sum.\\Sum.} & \makecell{Sum.\\Fall} & \makecell{Sum.\\Win.} & \makecell{Fall\\Fall} & \makecell{Fall\\Win.} & \makecell{Win.\\Win.} \\
\hline
Dates & \makecell{05-19\\05-22}  & \makecell{05-19\\08-13} & \makecell{05-19\\10-30} & \makecell{05-19\\02-10} & \makecell{08-13\\07-14} & \makecell{08-13\\10-30} &\makecell{ 08-13\\02-10} & \makecell{10-30\\11-28} & \makecell{10-30\\02-10} & \makecell{02-10\\12-12} \\
\hline
\end{tabular}
\caption{Test sequences on RobotCar dataset.}
\label{tb:pairs}
\end{table}
The RobotCar dataset is challenging for place recognition since visual appearance and brightness changes drastically. Snapshots of RobotCar are shown in Fig. \ref{fig:robotcar}. The testing pairs are given in Table~\ref{tb:pairs}, covering all combinations of seasons. 

\paragraph*{Accuracy}
Since the authors of \cite{ye2017place} have not published their code, we evaluate our algorithms using the same settings for fair comparisons. Specifically, we use the same segment as illustrated in Fig. 5(a) of \cite{ye2017place} for testing. When computing precision and recall, we also use $25$ meters as the GPS distance threshold. 

\begin{table}
\begin{subtable}[t]{\textwidth}
\centering
\tabcolsep=0.07cm
\begin{tabular}{|l|l|l|l|l|l|l|l|l|l|l|}
\hline
Tests & \makecell{Spr.\\Spr.} & \makecell{Spr.\\Sum.} & \makecell{Spr.\\Fall} & \makecell{Spr.\\Win.} & \makecell{Sum.\\Sum.} & \makecell{Sum.\\Fall} & \makecell{Sum.\\Win.} & \makecell{Fall\\Fall} & \makecell{Fall\\Win.} & \makecell{Win.\\Win.} \\
\hline
\cite{ye2017place}  & 0.774 & 0.736 & 0.589 & 0.419 & 0.764 & 0.557 & 0.489 & 0.599 & 0.443 & 0.597 \\
\hline
NBLD  & 0.651 & 0.700 & 0.611 & 0.351 & 0.672 & 0.496 & 0.379 & 0.454 & 0.351 & 0.491 \\
\hline
\hline
DELI. & 0.869 & 0.677 & 0.445 & 0.040 & 0.836 & 0.612 & 0.008 & 0.498 & 0.003 & 0.014 \\
\hline
M2DP  & 0.900 & 0.851 & 0.498 & 0.322 & 0.853 & 0.519 & 0.276 & 0.540 & 0.349 & 0.541 \\
\hline
S.C.  & \textbf{0.956} & \textbf{0.944} & \textbf{0.782} & 0.729 & \textbf{0.928} & \textbf{0.779} & 0.618 & \textbf{0.644} & 0.491 & \textbf{0.814} \\
\hline
\hline
BoW  & 0.558 & 0.342 & 0.208 & 0.300 & 0.305 & 0.418 & 0.371 & 0.002 & 0.293 & 0.001 \\
\hline
GIST  & 0.932 & 0.918 & 0.679 & \textbf{0.778} & 0.914 & 0.694 & \textbf{0.738} & 0.003 & \textbf{0.606} & 0.000 \\
\hline
\end{tabular}
\caption{AUC.}
\label{tb:auc}
\vspace{2mm}
\end{subtable}
\hspace{\fill}

\begin{subtable}[t]{\textwidth}
\centering
\tabcolsep=0.07cm
\vspace{2mm}
\begin{tabular}{|l|l|l|l|l|l|l|l|l|l|l|}
\hline
Tests & \makecell{Spr.\\Spr.} & \makecell{Spr.\\Sum.} & \makecell{Spr.\\Fall} & \makecell{Spr.\\Win.} & \makecell{Sum.\\Sum.} & \makecell{Sum.\\Fall} & \makecell{Sum.\\Win.} & \makecell{Fall\\Fall} & \makecell{Fall\\Win.} & \makecell{Win.\\Win.} \\
\hline
DELI. & 0.334 & 0.070 & 0.026 & 0.000 & 0.434 & 0.187 & 0.000 & 0.055 & 0.000 & 0.008 \\
\hline
M2DP  & 0.302 & 0.232 & 0.001 & 0.010 & 0.032 & 0.011 & 0.058 & 0.117 & 0.039 & 0.013 \\
\hline
S.C.  & 0.758 & \textbf{0.558} & \textbf{0.408} & \textbf{0.322} & \textbf{0.685} & \textbf{0.415} & \textbf{0.325} & \textbf{0.346} & \textbf{0.247} & \textbf{0.519} \\
\hline
\hline
BoW  & 0.032 & 0.021 & 0.023 & 0.031 & 0.005 & 0.034 & 0.100 & 0.000 & 0.043 & 0.000 \\
\hline
GIST  & \textbf{0.794} & 0.377 & 0.242 & 0.176 & 0.503 & 0.242 & 0.156 & 0.000 & 0.109 & 0.000 \\
\hline
\end{tabular}
\caption{Maximal recall at 100\% precision.}
\label{tb:recall}
\vspace{2mm}
\end{subtable}
\caption{Place recognition accuracy on RobotCar dataset.}
\end{table}

Table~\ref{tb:auc} shows the AUC of each algorithm running the tests in Table~\ref{tb:pairs}. Since we cannot re-run the tests in \cite{ye2017place}, data of \cite{ye2017place} and NBLD are taken directly from \cite{ye2017place}, which are just for reference; whereas the rows marked ``\verb|DELI.|'', ``\verb|M2DP|'', and ``\verb|S.C.|'' represent our approaches adapting these three global descriptors. Table~\ref{tb:recall} illustrates the maximal recall with 100\% precision. Scan Context (``\verb|S.C.|'') achieves both the highest AUC and the highest recall in most tests. Scan Context depends on the height range for place recognition. In the RobotCar dataset, the maximum/minimum height is usually from buildings/ground. Therefore, the height range is not very sensitive to seasonal change. GIST behaves best in the rest of the tests. It works well because the viewing angle is mostly unchanged in the RobotCar dataset. M2DP is the next-best performing approach, which has a relatively high accuracy when there is less visual appearance change (\eg \textit{Test a}: Spring-Spring). However, in different seasons, the trees along the streets have vastly different appearances (\eg for losing leaves) as illustrated in Fig. \ref{fig:robotcar}. M2DP projects all 3D points to get a signature, so its accuracy drops. DELIGHT suffers more from visual appearance change between seasons because it purely depends on grayscale intensity, which indicates the importance of 3D structure for place recognition. BoW has the worst performance; a potential reason is that feature matching is sensitive to changing scene factors such as trees and traffic (vehicular and pedestrian).

\begin{figure}
  \centering
    \includegraphics[width=\textwidth]{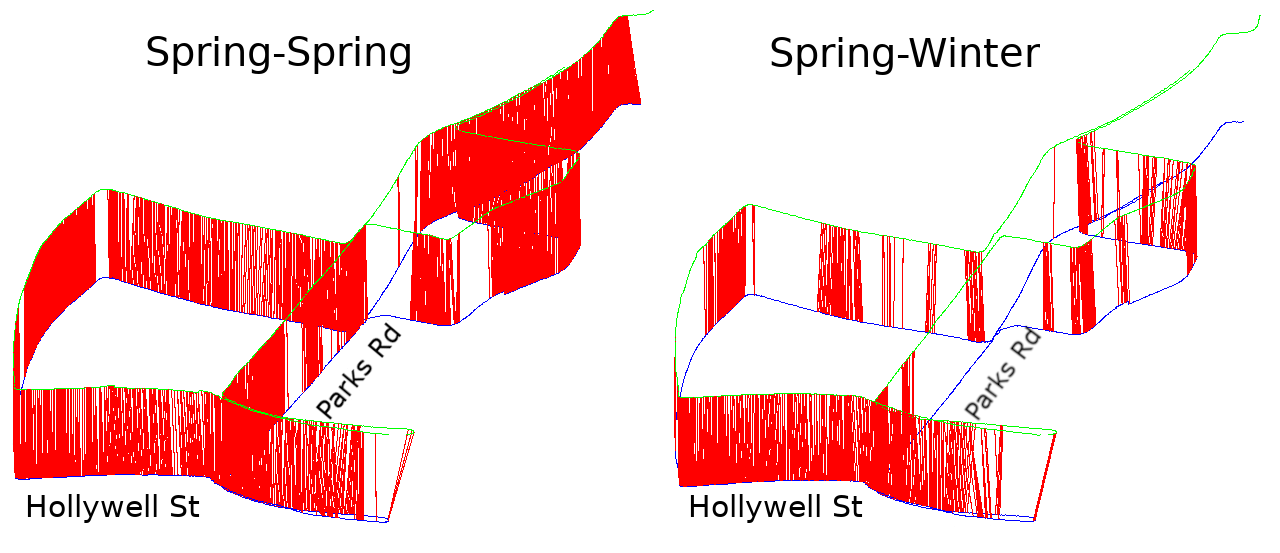}
  \caption{Places recognized (marked as red) by Scan Context at 100\% precision.}
  \label{fig:traj_rd}
\end{figure}

Fig. \ref{fig:traj_rd} shows the places recognized by Scan Context at 100\% precision. For the easy case (Spring-Spring), Scan Context recognizes most of the places. For the challenging case (Spring-Winter), with the reference of Fig. \ref{fig:robotcar}, most places along Holywell street are correctly recognized since it is occupied mostly with buildings; but most places along Parks road are not recognized because there are many trees (maximum height) on both sides of the road. 

\begin{figure}
  \centering
    \includegraphics[width=\textwidth]{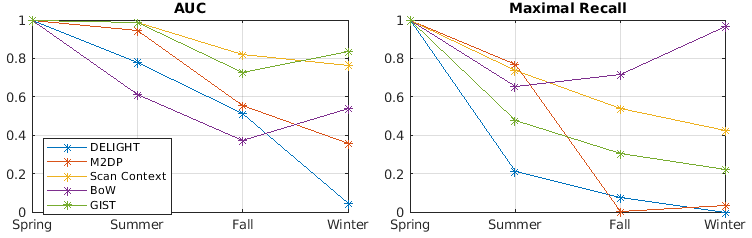}
  \caption{Robustness against seasonal visual appearance change, using spring as query season. Values are normalized by Spring-Spring.}
  \label{fig:sensitivity}
\end{figure}

Fig. \ref{fig:sensitivity} shows the decrease in accuracy of each algorithm against seasonal visual appearance change. For AUC, Scan Context and GIST outperform the rest in terms of robustness; for maximal recall, BoW has an abnormal curve because all of its maximal recall values are very low (Table \ref{tb:recall}). Other than that, Scan Context is the most robust one. Therefore, we conclude that Scan Context is robust against visual appearance change.

\begin{figure}
\centering
\includegraphics[width=\textwidth]{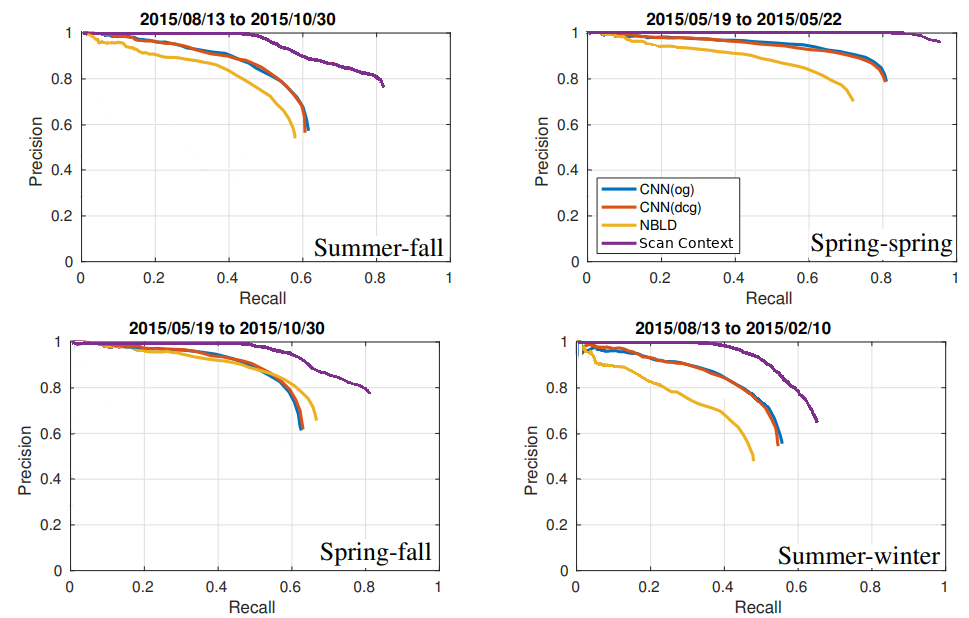}
\caption{Precision-recall curves of Scan Context compared with that of \cite{ye2017place}.}
\label{fig:pr-curve}
\end{figure}

Fig. \ref{fig:pr-curve} shows the comparison of the precision-recall curve between the adapted Scan Context and the methods in \cite{ye2017place}. It validates that Scan Context adapted in this work outperforms \cite{ye2017place} in accuracy. 

\begin{table}
\centering
\tabcolsep=0.06cm
\begin{tabular}{|l|l|l|l|l|l|l|l|l|l|l|}
\hline
Tests & \makecell{Spr.\\Spr.} & \makecell{Spr.\\Sum.} & \makecell{Spr.\\Fall} & \makecell{Spr.\\Win.} & \makecell{Sum.\\Sum.} & \makecell{Sum.\\Fall} & \makecell{Sum.\\Win.} & \makecell{Fall\\Fall} & \makecell{Fall\\Win.} & \makecell{Win.\\Win.} \\
\hline
Structure & \makecell{0.955\\0.270 } & \makecell{0.940\\0.216 } & \makecell{0.762\\0.390 } & \makecell{0.699\\0.154 } & \makecell{\textbf{0.931}\\0.279 } & \makecell{0.753\\0.066 } & \makecell{0.610\\0.105 } & \makecell{\textbf{0.652}\\0.147 } & \makecell{\textbf{0.500}\\0.049 } & \makecell{0.778\\0.134 } \\
\hline
Intensity & \makecell{0.834\\0.230 } & \makecell{0.645\\0.050 } & \makecell{0.344\\0.039 } & \makecell{0.112\\0.021 } & \makecell{0.831\\0.151 } & \makecell{0.390\\0.057 } & \makecell{0.086\\0.027 } & \makecell{0.290\\0.056 } & \makecell{0.096\\0.027 } & \makecell{0.478\\0.032 } \\
\hline
Fused & \makecell{\textbf{0.956}\\\textbf{0.758} } & \makecell{\textbf{0.944}\\\textbf{0.558} } & \makecell{\textbf{0.782}\\\textbf{0.408} } & \makecell{\textbf{0.729}\\\textbf{0.322} } & \makecell{0.928\\\textbf{0.685} } & \makecell{\textbf{0.779}\\\textbf{0.415} } & \makecell{\textbf{0.681}\\\textbf{0.325} } & \makecell{0.644\\\textbf{0.346} } & \makecell{0.491\\\textbf{0.247} } & \makecell{\textbf{0.814}\\\textbf{0.519} } \\
\hline
\end{tabular}
\caption{AUC (top sub-rows) and maximal recall (bottom sub-rows) at 100\% precision of Scan Context with structure and/or grayscale intensity.}
\label{tb:intensity}
\end{table}

\paragraph*{Intensity Contribution} Table~\ref{tb:intensity} shows the AUC and maximal recall of Scan Context using structure only, intensity only, and both structure and intensity. Scan Context with intensity only performs poorly on the RobotCar dataset, since grayscale intensity changes drastically throughout different seasons. However, augmenting the structure descriptor with intensity information clearly improves the maximal recall, even though it does not obviously improve the AUC.

We do not include the efficiency comparison on the RobotCar dataset since we use the identical setup with the KITTI dataset and the results are similar. 

\subsection*{Use Case Analysis}
After the experiments, we claim that the use case of the proposed method is for forward-moving vehicles (for accumulating points) equipped with stereo cameras in visually changing environments (\eg RobotCar dataset), where the proposed approach recognizes place with high accuracy, efficiency, and robustness. It also works with repetitive texture (\eg Sequence 02 of KITTI dataset). Additionally, adopting the proposed approach is easier than adopting BoW for direct vSLAM. 

The conventional BoW works on individual images, and there is no forward-moving constraint. It achieves higher accuracy, especially for small loop segments (\eg Fig. \ref{fig:kitti-plot}) when there is not much visual appearance change. 

\section{Conclusions}
In this paper, we propose a novel place recognition approach for stereo-visual odometry. Instead of 2D image similarity, we depend on the 3D points generated by the visual odometry to determine the correlation between places. The 3D points from stereo systems with an absolute scale are used to imitate LiDAR scans which are fed into three global LiDAR descriptors, which are DELIGHT, M2DP, and Scan Context. We augment the descriptors with grayscale intensity information. Experiments on the KITTI dataset and RobotCar dataset show the accuracy, efficiency, and robustness of the proposed method.

For the next step, we will integrate the proposed method into state-of-the-art stereo-visual odometry algorithms for loop closure detection, and quantify the performance improvement in visually challenging environments. Furthermore, we intend to extend the proposed approach with deep learning and compare to learning-based place recognition approaches.

\section*{Acknowledgment}
This work was partially supported by the Minnesota Robotics Institute Seed (MnRI) Grant. 

\bibliographystyle{plain}
\bibliography{main}
\end{document}